\RequirePackage{fix-cm}
\documentclass[twocolumn]{svjour3}          % twocolumn
\smartqed  % flush right qed marks, e.g. at end of proof
\usepackage{amsmath,amsfonts}
\usepackage{algorithmic}
\usepackage{algorithm}
\usepackage{array}
\usepackage[caption=false,font=normalsize,labelfont=sf,textfont=sf]{subfig}
\usepackage{textcomp}
\usepackage{stfloats}
\usepackage{url}
\usepackage{verbatim}
\usepackage{graphicx}
\usepackage{cite}
\usepackage{multirow}
\usepackage{color}
\usepackage{booktabs}

\begin{document}
\begin{sloppypar}
\title{Multi-Weather Image Restoration via Histogram-Based 
Transformer Feature Enhancement}

\author{Yang Wen\textsuperscript{a} \and
Anyu Lai\textsuperscript{a} \and
Bo Qian\textsuperscript{b} \and Hao Wang\textsuperscript{a} \and Wuzhen Shi\textsuperscript{a,*} \and Wenming Cao\textsuperscript{a}}

\institute{ \textsuperscript{a} Yang Wen, Anyu Lai, Hao Wang Wuzhen Shi and Wenming Cao are with the Guangdong Provincial Key Laboratory of Intelligent Information Processing, School of Electronic and Information Engineering, Shenzhen University, Shenzhen, China. 
\\\textsuperscript{b} Bo Qian is with the Department of Comupter Science and Engineering, Shanghai Jiao Tong University, Shanghai, China.}
 
%\subtitle{Insert your subtitle here, otherwise leave blank}
%\author{Jixue Tang\textsuperscript{1}  \and Yang Wen\textsuperscript{\Letter}  }
%\institute{Jixue Tang \at Department of Comupter Science and Engineering, Shanghai Jiao Tong University, Shanghai, China (email: tang.jixue@sjtu.edu.cn)
%\and Yang Wen \at School of Electronic and Information Engineering, Shenzhen University, Shenzhen, China
%}
% \and Dagan Feng \at School of Computer Science, The University of Sydney, Sydney, Australia \at  Med-X Research Institute. Shanghai Jiao Tong University, Shanghai, China}
\date{ }% The correct dates will be entered by the editor

\maketitle

%\footnotetext[1]{Jixue Tang and Xiaolong Yang contributed equally to this work.}
\footnotetext[1]{Corresponding author: Wuzhen Shi(email: wzhshi@szu.edu.cn).}
\footnotetext[2]{This work was supported in part by the National Science Foundation of China under Grants 62301330 and 62101346, and in part by the Guangdong Basic and Applied Basic Research Foundation under Grants 2024A1515010496, 2021A1515011702 and 2022A1515110101.}

\begin{abstract}
Currently, the mainstream restoration tasks under adverse weather conditions have predominantly focused on single-weather scenarios. However, in reality, multiple weather conditions always coexist and their degree of mixing is usually unknown. Under such complex and diverse weather conditions, single-weather restoration models struggle to meet practical demands. This is particularly critical in fields such as autonomous driving, where there is an urgent need for a model capable of effectively handling mixed weather conditions and enhancing image quality in an automated manner. In this paper, we propose a Task Sequence Generator module that, in conjunction with the Task Intra-patch Block, effectively extracts task-specific features embedded in degraded images. The Task Intra-patch Block introduces an external learnable sequence that aids the network in capturing task-specific information. Additionally, we employ a histogram-based transformer module as the backbone of our network, enabling the capture of both global and local dynamic range features. Our proposed model achieves state-of-the-art performance on public datasets.

\end{abstract}

\section{Introduction}
\label{sec:introduction}
The removal of adverse weather conditions, including rain, snow, and haze, from images is a critical challenge in numerous fields. Extreme weather events significantly impair the ability of computer vision algorithms to extract relevant information from images, necessitating the mitigation of such weather-related effects to enhance the reliability of computer vision systems\cite{Du_Xu_Zhen_Cheng_Shao_2020}\cite{Ren_Zuo_Hu_Zhu_Meng_2019}\cite{Yang_Tan_Wang_Fang_Liu_2021}.

\begin{figure}[t]
    \begin{center}
    \includegraphics [width=1\columnwidth]{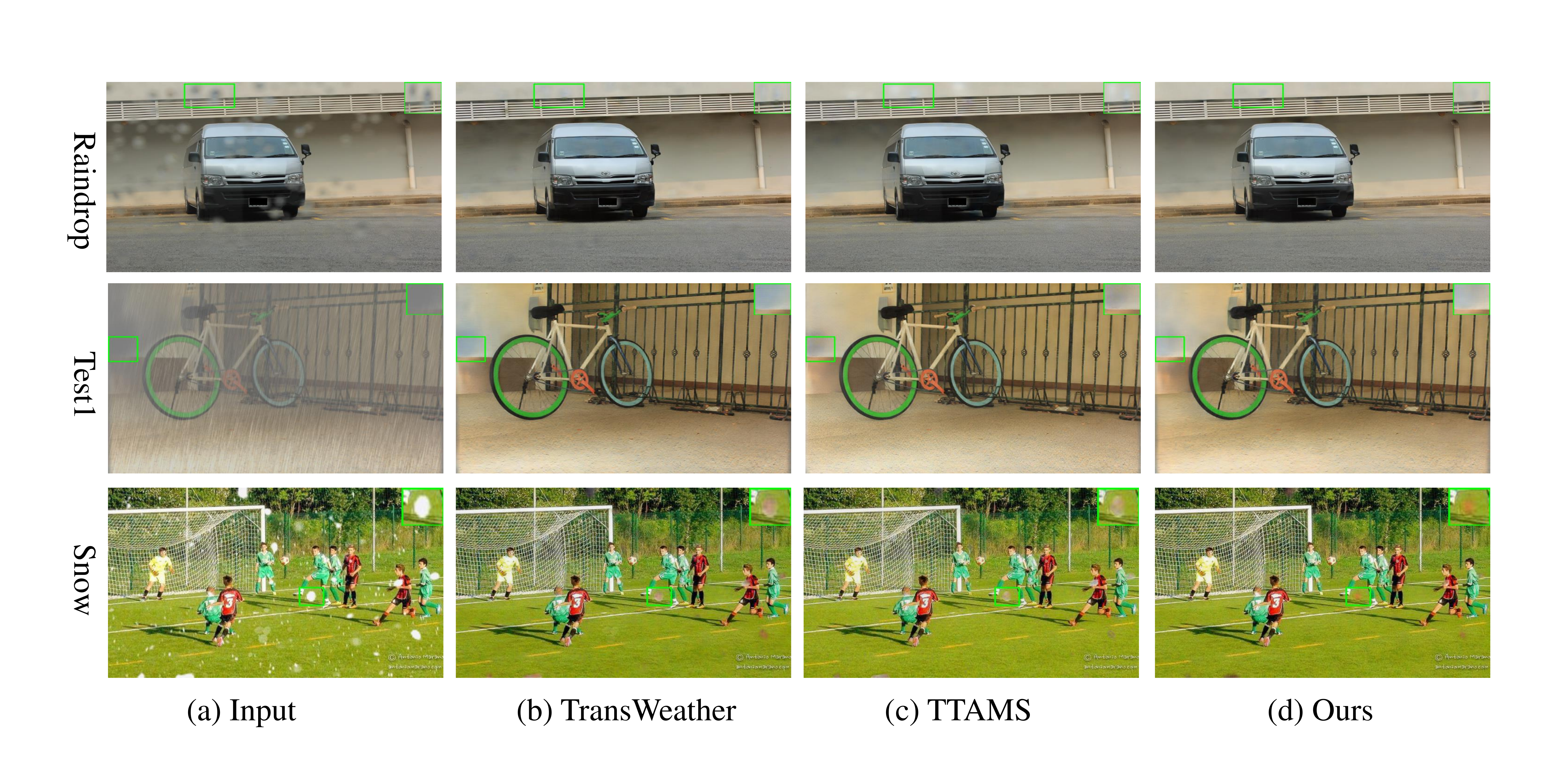}
    \end{center}
    \caption{The proposed method and example results of TransWeather\cite{JeyaMariaJoseValanarasu2022TransWeatherTR} and TTAMS\cite{wen2024multipleweatherimagesrestoration} for multi-weather image restoration.}
    \label{figure1}
    \end{figure}

Normally, the formation of hazy can be defined as:

\begin{equation}
    I(x)=J(x)t(x)+A(1-t(x))
\end{equation}

where $I(x)$ and $J(x)$ are the radiance of the degraded hazy image and the target hazy-free scene, respectively. $A(\cdot)$ is the global atmospheric light, and $t(x)$ is the medium transmission map, which depends on the unknown depth information. Most previous dehazing methods\cite{Berman_Treibitz_Avidan_2016}\cite{Hautiere_Tarel_Aubert_2007}first estimate the transmission map $t(x)$ or atmospheric light $A$, and then try to recover the final clean image $J(x)$.
However, methods based on prior knowledge tend to perform poorly in some situations. With the development of deep learning, many methods based on CNN have been used in dehazing\cite{Luo_Chang_Bo_2022}\cite{Ren_Liu_Zhang_Pan_Cao_Yang_2016}. Luo et al.\cite{Luo_Chang_Bo_2022} proposed an end-to-end multi-feature fusion network for single image dehazing. Ren et al.\cite{Ren_Liu_Zhang_Pan_Cao_Yang_2016} propose a multi-scale deep neural network for single image dehazing by learning the mapping between a hazy image and its transmission map. In literature\cite{Li_Wu_Lin_Liu_Zha_2018}, the degradation of rain marks is defined as:

\begin{equation}
    I=J+\sum_{i}^{n}S_i 
\end{equation}

where $I$ is the observed input image. $J$ is the background scene without rain. $S_i$ is the rain layer, $n$ is the total number of rain striae. Traditional unsupervised (i.e., prior-based) methods in the context of rain removal aim to exploit the inherent structure of background and rain layers as prior knowledge to effectively constrain the solution space of a carefully designed optimization model. These methods commonly leverage frequency information and sparse representations as prevalent priors\cite{Zhang_Patel_2017}\cite{Zhu_Fu_Lischinski_Heng_2017}\cite{Gu_Meng_Zuo_Zhang_2017}. Zhang et al.\cite{Zhang_Patel_2017} first learn a general set of sparsity-based and low-rank representation based convolutional filters for efficiently representing background-clear images and rain streaks, respectively.  Li et al.\cite{Li-Wei_Kang_Chia-Wen_Lin_Yu-Hsiang_Fu_2012} proposed a method to separate the rain layer from the rainy image and derain the image layer to remove the visual impact of rain in a single rainy image.
Other approaches utilize the strong non-linear modeling capability of deep learning, applying convolutional neural networks\cite{Fu_Huang_Zeng_Huang_Ding_Paisley_2017}\cite{Fu_Huang_Ding_Liao_Paisley_2017}, adversarial networks\cite{Zhang_Sindagi_Patel_2020}\cite{Zhang_Patel_2018}, recurrent and multi-stage networks\cite{Ren_Zuo_Hu_Zhu_Meng_2019}\cite{Li_Wu_Lin_Liu_Zha_2018} to restore images. According to previous studies\cite{Chen_Fang_Ding_Tsai_Kuo_2020}\cite{Chen_Fang_Hsieh_Tsai_Chen_Ding_Kuo_2021}, snow degradation can be modeled as:
\begin{equation}
    I(x)=K(x)T(x)+A(x)(1-T(x))
\end{equation}

where $I(x)$ denotes the snow image, $T(x)$ and $A(x)$ denote the transmission map and atmospheric light, respectively. $K(x)$, the unobstructed snow scene, can be decomposed into $K(x) = J (x)(1-Z(x)R(x)) + C(x)Z(x)R(x)$, where $J(x)$ is a clean image, and $R(x)$ is a binary mask representing the location information of snow. $Z(x)$ and $C(x)$ represent the color difference image and snow mask. Currently, snow removal methods are mainly divided into two categories: prior knowledge-based and deep learning-based methods. There are methods based on prior knowledge such as using color features\cite{Soo-Chang_Pei_Yu-Tai_Tsai_Chen-Yu_Lee_2014}, based on guided images\cite{Xu_Zhao_Liu_Tang_2012}, and based on multi-guided filter feature extraction\cite{Zheng_Liao_Guo_Fu_Ding_2013}. 
With the rise of deep learning, several CNN-based methods\cite{Li_Zhang_Fang_Huang_Jiang_Gao_Hwang_2019}\cite{Chen_Fang_Ding_Tsai_Kuo_2020}\cite{Liu_Jaw_Huang_Hwang_2018} have been proposed. These methods are based on modeling a single task, and deploying such models in real-world applications is challenging because it requires the development and deployment of a large number of specialized models to handle different weather conditions, resulting in complex and bulky systems. This paper reviews the state-of-the-art in CNN-based multi-task weather-related image restoration methods, aiming to develop a unified model capable of handling multiple degradation types. The effectiveness of these methods is evaluated using various benchmarks and their potential applications in real-world scenarios are discussed.

Recently Li et al\cite{RuotengLi2020AllIO}. proposed an innovative approach to weather-related image restoration, which utilizes an All-in-One network to handle multiple degradation scenarios using a single model. The model incorporates different encoders for various weather conditions and a unified decoder to restore clear images. However, the effectiveness of the model may be limited in removing degradation from local details, and the use of multiple encoders results in higher model complexity. Valanarasu et al. proposed the TransWeather network\cite{JeyaMariaJoseValanarasu2022TransWeatherTR}, which embeds weather-type information into the network and employs intra-patch transformer (Intra-PT) blocks to extract fine detail features for low-level vision tasks like weather removal. However, the model may encounter challenges in addressing large-scale weather degradation scenarios due to its lack of attention to global background information. When confronted with the degradation of extensive patches, the model tends to extract information that includes the degradation itself, posing difficulties in effectively restoring the image. However, when faced with a substantial degradation area, it becomes crucial to extract information from the broader global background in order to guide the image restoration process.

In this paper, we propose a novel approach for restoring images degraded by various weather conditions, utilizing a Task Sequence Generator and a histogram-based Transformer module. During the feature extraction phase, we employ a Task Intra-patch Block (TIPB) to divide the image into smaller patches, from which degradation features are extracted. These features are not only used in subsequent feature extraction but also input into the Task-Sequence Generator, which generates task sequences based on the input task features at each stage. Inspired by Li et al\cite{sun2024restoring}, in the main network, the histogram-based Transformer is used to distinguish between degradation information and background information, facilitating more effective removal of weather-induced degradation. To better integrate the features between the backbone network and the task sequences, we use a Cross-Self-Attention module to interactively process the information extracted by the two modules. Finally, to blend the degradation and background information during image restoration, we adopt an adaptive upsampling technique. The sample results obtained with the proposed method are displayed in Figure\ref{figure1}. Our main contributions are:

\begin{itemize}
    \item We introduce a novel and efficient solution to tackle the challenge of adverse weather removal, with a particular focus on image restoration guided by weather degradation information and task feature sequence generation. Our proposed method surpasses the performance of existing state-of-the-art approaches on real-world datasets and downstream object detection tasks. 
    \item We propose the Task Sequence Generator, a novel feature extraction block designed to effectively capture detailed features of various degradation types across multiple scales. The multi-scale degradation types are fed into the Task Sequence Generator to extract task-specific features, enabling our method to generate rich, stage-specific features tailored for each phase of the image restoration process. This approach allows us to effectively address different types of degradation, achieving exceptional performance in restoring degraded images. 
    \item We employed a Histogram Transformer Block which provides spatial attention to the dynamic range of weather-induced degradation. This enables globally effective degradation removal by focusing on relevant regions across the entire image. 
\end{itemize}

\section{Related work}
\label{sec:related work}

Deep learning-based solutions have become increasingly popular for various weather-related image restoration tasks, including rain removal\cite{QingGuo2020EfficientDeRainLP}\cite{KuiJiang2020MultiScalePF}, hazy removal\cite{YeyingJin2022StructureRN}\cite{WeiLiu2022HolisticAA}, and snow removal\cite{YunFuLiu2017DesnowNetCD}\cite{WeiTingChen2021ALLSR}. These approaches have demonstrated significant performance improvements compared to traditional methods.

\subsection{For Rain Removal} Guo et al.\cite{QingGuo2020EfficientDeRainLP} were the first to introduce an innovative pixel-wise dilated filtering technique. Specifically, they filtered rainy weather images using a pixel-level kernel, which is estimated from a kernel prediction network. This network efficiently predicts an appropriate multi-scale kernel for each pixel. To address the gap between synthetic and real data, they also proposed an effective data augmentation method that enhances the network's ability to process real rainy images. Kui et al.\cite{KuiJiang2020MultiScalePF} primarily proposed a Multi-Scale Progressive Fusion Network (MSPFN). Considering the imaging principles of rain, where the distance between the rain and the camera varies, rain in images appears with different degrees of blurriness and resolution. The complementary information across different resolutions and pixels can thus be leveraged to represent rain streaks. The main contribution of this paper lies in proposing a framework based on input geometry and depth graphics, exploring the multi-geometry representation of rain streaks, and achieving effective deraining. By calculating gradients for rain streaks at different locations, they obtain global texture information, which is used to explore spatial complementarity and to extract information that characterizes the rain streaks. Jiang et al.\cite{Jiang_Wang_Yi_Chen_Huang_Luo_Ma_Jiang_2020} employed cyclic calculations to capture global patterns of rain streaks at various locations, thereby reducing redundant information in the spatial dimension. They also utilized a pyramid structure to introduce an attention mechanism, guiding the fine fusion of relevant information across different scales. Ren et al.\cite{Ren_Zuo_Hu_Zhu_Meng_2019} proposed using a repeatedly expanded ResNet to leverage deep image features across stages to predict the final output.

\subsection{For Haze Removal} To tackle the challenge of dense and uneven haze distribution, Jin et al.\cite{YeyingJin2022StructureRN} proposed a model that leverages feature representations extracted from a pre-trained visual transformer (DINO-ViT) to restore background details. To enhance the network's focus on non-uniform haze regions and effectively remove haze, they introduced an uncertainty feedback learning mechanism. This mechanism generates uncertainty maps that exhibit higher uncertainty in denser haze regions, effectively serving as attention maps that represent haze density and distribution irregularities. The feedback network iteratively refines the dehazing output using these uncertainty maps. Liu et al.\cite{WeiLiu2022HolisticAA} developed an innovative generative adversarial network named the Holistic Attention Fusion Adversarial Network (HAAN). This network is designed to learn the holistic channel-spatial feature correlations, effectively exploiting the self-similarity of texture and structural information. To preserve vibrant colors and rich textural details, they incorporated an atmospheric scattering model within the hazy synthesis module, guiding the generation process through a novel sky segmentation network for atmospheric light optimization. Mo et al.\cite{Mo_Li_Ren_Shang_Wang_Wu_2022} designed a prior map that inverts the dark channel and employs max-min normalization to suppress sky regions while emphasizing objects, thereby enhancing the restoration of hazy images, particularly over water surfaces. Zhang et al.\cite{Zhang_Hou_2017} addressed the color distortion problem associated with dehazing algorithms based on the dark channel prior. By considering the effect of incident light frequency on each color channel, they derived the ratio between the transmittance of each channel. This method effectively restores clear images with high color saturation by adjusting the transmittance ratios appropriately.

 \subsection{For Snow Removal} Handcrafted features remain the dominant approach for snow removal, limiting the potential for large-scale generalization. Liu et al.\cite{YunFuLiu2017DesnowNetCD} addressed this challenge by designing a multi-stage network called DesnowNet, which sequentially removes translucent and opaque snow particles. Their method distinguishes between the translucency and color differences of snow to achieve accurate estimation. Additionally, they separately estimate the remaining components of snow-free images to recover details obscured by opaque snow. The entire network incorporates a multi-scale design to simulate the diverse appearance of snow. Chen et al.\cite{WeiTingChen2021ALLSR} proposed a single-image snow removal algorithm that accounts for the diversity in snow particle shapes and sizes. To better represent the complexity of snow shapes, they employed dual-tree wavelet transform and introduced a complex wavelet loss within the network. Furthermore, they proposed a hierarchical decomposition paradigm to effectively handle snow particles of varying sizes. Their method also introduced a novel feature known as the Contradictory Channel (CC) for snow scenes. Sixiang Chen et al.\cite{Chen_Ye_Liu_Chen_2022} utilized cross-attention mechanisms to establish local-global contextual interactions between scale-aware snow queries and local patches, effectively tackling the challenge of single-image snow removal. Ye et al.\cite{Ye_Chen_Liu_Chen_Li_2022} proposed an efficient and compact recovery network, constructed using multiple expert systems and an adaptive gating network. This network extracts degradation information from the input image and adaptively modulates the output of task-specific expert networks to mitigate the effects of winter weather.

\subsection{Multi-Task Weather-Related Image Restoration} The aforementioned approaches are predominantly centered on a singular task in modeling, which makes them challenging to employ in practical scenarios due to the diverse range of weather conditions encountered in real-life situations. Consequently, our model requires the capability to effectively adapt and handle a multitude of weather conditions at any given moment.
Li et al.\cite{RuotengLi2020AllIO} first designed a generator with multiple task-specific encoders, each associated with a specific type of severe weather degradation. They utilize a neural architecture search to optimally process image features extracted from all encoders. Subsequently, to transform degraded image features into clean background features, they introduce a series of tensor-based operations that encapsulate the fundamental physics behind the formation of rain, haze, snow, and adherent raindrops. These operations are the basic building blocks of schema search. Finally, the discriminator simultaneously evaluates the correctness of the restored image and classifies the degradation type. Valanarasu et al.\cite{JeyaMariaJoseValanarasu2022TransWeatherTR} propose TransWeather, a Transformer-based end-to-end model that can recover images degraded by any weather condition with only one encoder and one decoder. Specifically, they exploit a novel Transformer encoder that uses intra-patch Transformer blocks to enhance intra-patch attention to effectively remove small weather degradations. We also introduce a transformer decoder with learnable weather-type embeddings to adapt to current weather degradations.

The aforementioned approach, Transweather, suffers from certain limitations, notably the disregarding of global information in image restoration, leading to suboptimal outcomes when restoring images with large-scale degradation. Additionally, the All-in-One network employs different encoders tailored for specific tasks, rendering it challenging to deploy. Our proposed network addresses these limitations by incorporating a versatile encoder and decoder for various tasks, while simultaneously enhancing the network's receptive field through FFC, enabling it to account for the global context when restoring images with extensive degradation.

\section{Method}
\label{sec:method}
\begin{figure*}
\begin{center}  
    \includegraphics [width=1\textwidth]{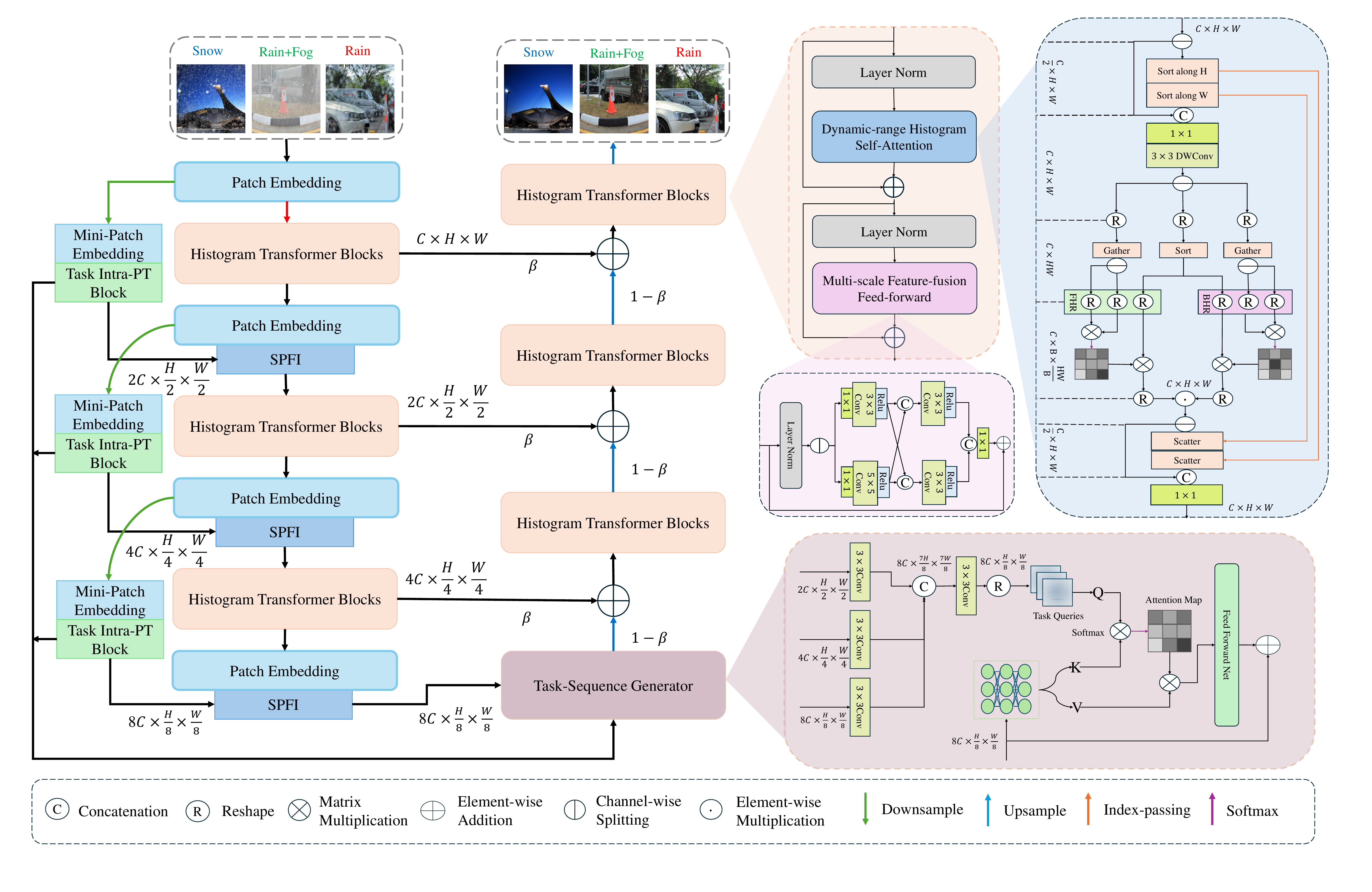}
    \end{center}
\caption{Overview of the proposed network. The degraded image is progressively processed through Histogram Transformer Blocks and Task Intra-PT Blocks. The degradation features and task features obtained from these blocks are fused using the SPFI block. The multi-level task features are then passed through the Task Sequence Generator to produce a task sequence. Finally, the clean image is restored through adaptive upsampling. }
\label{fig_2}
\end{figure*}

We propose a novel framework to tackle different image degradation tasks, as shown in Figure\ref{fig_2}. In this section, we provide a comprehensive overview of the network framework, including a detailed description of each individual module that comprises the network.

% \begin{list}{}{}
% \item{\url{http://www.latex-community.org/}} 
% \item{\url{https://tex.stackexchange.com/} }
% \end{list}

\subsection{Network Architecture}
The proposed network takes 3×3 weather-degraded images as input and uses a multi-level histogram-based Transformer module to capture dynamic range attention related to degradation at different scales. The output of each stage is fed into the Task-In-Patch Block (TIPB), which extracts specific degradation details at a finer scale. These details are then merged with the next level's histogram-based Transformer module through a cross-self-attention mechanism, effectively integrating task features with degradation features. The task features generated at different stages are input into the Task Sequence Generator to produce a specific task sequence, aiding in the identification of the particular degradation affecting the input image. To better handle the relationship between task features and image information, learnable parameters are utilized during the upsampling stage to selectively fuse task features with image features, ultimately restoring a clear image. 

\subsection{Task Intra-Patch Block}
\begin{figure}[H]
    \begin{center}
    \includegraphics [width=1\columnwidth]{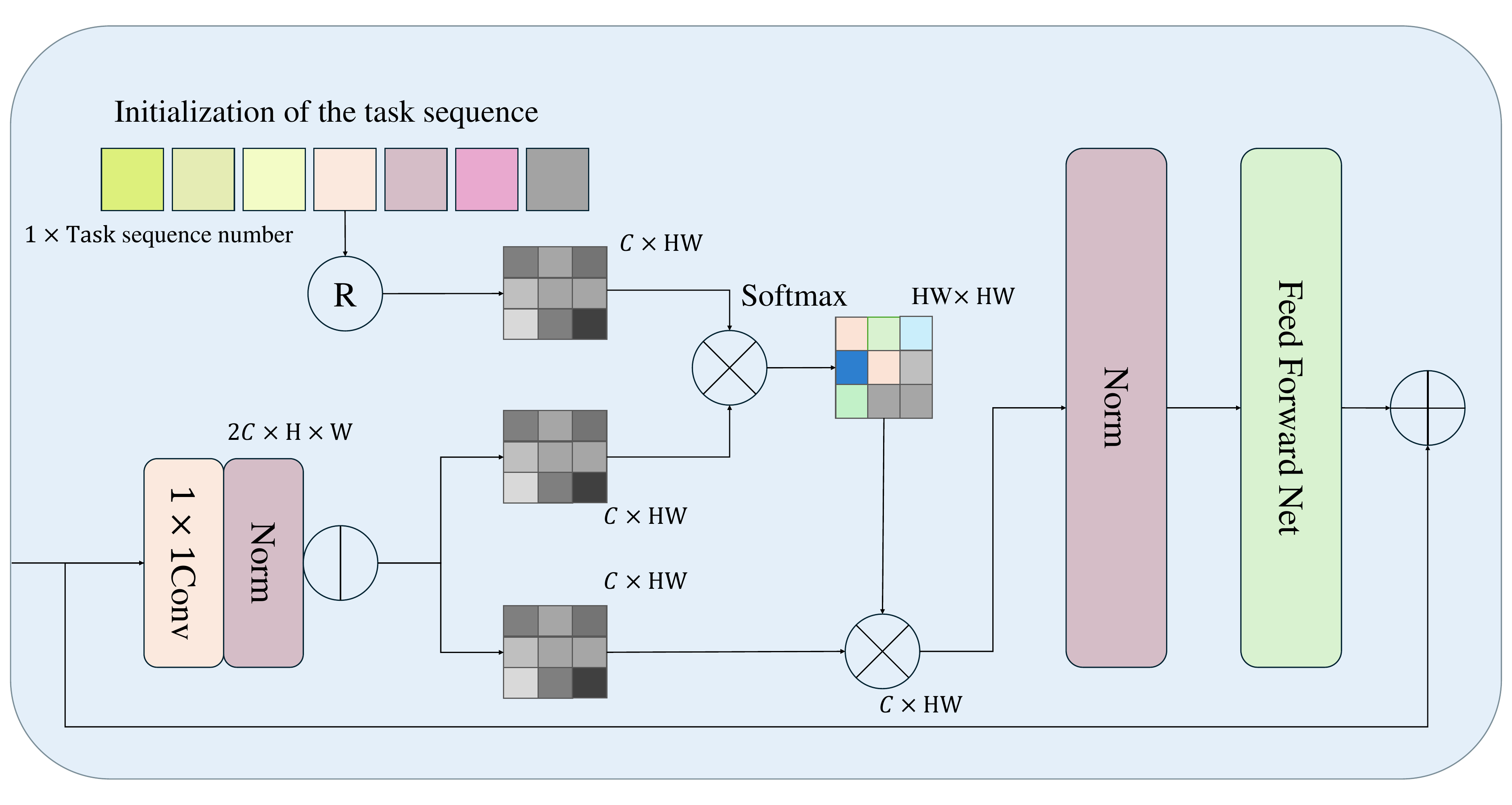}
    \end{center}
    \caption{Detail of Task Transformer Block. By calculating the q introduced from the outside and the kv generated by the image, the attention map is input to the multi-layer perceptron to obtain the feature map with task information.}
    \label{figure3}
    \end{figure}

At each stage, the Task Intra-patch Block (TIPB) processes the image features by first cropping them to half the size of the original image, enabling the extraction of finer degradation details. As illustrated in Figure \ref{figure3}, to adaptively query different degraded features, an external learnable sequence is introduced and optimized during the network's training phase. This sequence generates a feature map rich in task-specific information, which is then combined with the input image and fed into the Transform Block of the subsequent stage. The feature maps from all stages are collectively input into the Task Sequence Generator, which produces a task query vector that helps identify the specific degradation affecting the input image. This approach enables the effective extraction of task-specific information at each stage, ultimately enhancing image restoration performance. The output of TIPB can be expressed as:

\begin{equation}
    TIPB_{i}(I_i)=FFN(MSA(I_i)+I_i)
\end{equation}

where $T(\cdot)$ represents the transformer block, $FFN(\cdot)$ repre-
sents the feed-forward network block, $MSA(\cdot)$ represents
multi-head self-attention, $I$ is the input and $i$ represents the
stage in the encoder. The multi-head attention of the TIPB module is different from the traditional form, and its self-attention is defined as follows: 
\begin{equation}
    Attn(Q,K,V) = softmax(\frac{Q_{learnabled}K^T}{\sqrt{d}})V
\end{equation}
The proposed network leverages a randomly generated task query sequence ($Q$) to represent a diverse range of weather conditions. The keys ($K$) and values ($V$) used in the attention mechanism are derived from the input feature map.

\subsection{Task Sequence Generator}

The TIPB module introduces a stochastic task vector into the transformer module as a query within the attention mechanism. This vector, trained alongside the network, facilitates the capture of degradation characteristics under varying weather conditions. TIPB operates at each level of the encoder to extract degradation information across different scales in the image. The output of the TIPB is then passed to the encoder of the subsequent stage, with the outputs from all stages collectively serving as input to the Task Sequence Generator, which generates a task sequence relevant to the image.

The Task Sequence Generator consists of multiple convolutional layers of varying scales and a self-attention module. These convolutional layers, applied at different stages, enable efficient processing of the output from the Task Intra-Patch Block (TIPB). A 3x3 convolutional layer is then employed to merge task information from four different scales, resulting in a task feature query vector map. This map is used as the query (Q) in conjunction with the image within the self-attention mechanism, generating a feature map enriched with task-specific information. The output of Task sequence generator can be expressed as:
\textbf{\begin{equation}
    Tsg(I, Q_{Task})=FFN(MSA(I, Q_{Task})+I)
\end{equation}}

\textbf{\begin{equation}
   \small Q_{Task}\!\!=\!\!Cov_{3,3}(Cov_{7,7}(T_1)\!\!+\!\!Cov_{5,5}(T_2)\!\!+\!\!Cov_{3,3}(T_3))
\end{equation}}
Where $Tsg(\cdot)$ represents the output of the Task-sequence Generator, $FFN(\cdot)$, and $MSA(\cdot)$ represent the feedforward network and the multi-head self-attention module, respectively. I represent the feature map input to the Task-sequence Generator, and $Q_{Task}$ represents the generated task query sequence. $T_i$ denotes the output of TIPB from the i-th stage. $Conv_{n,n}$ represents the use of n×n convolution operations.

Our results demonstrate that the Task-sequence Generator module significantly improves the model's ability to capture degradation characteristics compared to the Base Model. In particular, the degradation information in the output images is clearer, and the contrast between the degraded content and the background is more pronounced. These findings underscore the importance of incorporating advanced techniques, such as the Task-sequence Generator module, in image restoration tasks to enhance performance and improve the quality of the results.

\subsection{Histogram Transformer Block}
The Histogram Transformer Block primarily consists of two components: the Dynamic-range Histogram Self-Attention (DHSA) and the Multi-scale Feature-fusion Feed-forward  (MSFF). These two components interact with layer normalization to form the complete Histogram Transformer Block, which is defined as follows:
\begin{equation}
    F_{l} = F_{l-1}+DHSA(LN(F_{l-1}))
\end{equation}

\begin{equation}
    F_{l} = F_{l}+MSFF(LN(F_{l}))
\end{equation}
where $F_{l}$ represents the feature at $l-th$ stage and $FN(\cdot)$ denotes layer normalization.Next, we will provide a detailed explanation of the implementation details of DHSA and  MSFF components.

To better capture the dynamically distributed weather-induced degradation, the DHSA module introduces a dynamic range histogram specifically designed for image restoration under adverse weather conditions. This module consists of a process involving dynamic range convolution, which reorders the spatial distribution of fractional features, and a dual-path histogram self-attention mechanism that integrates both global and local dynamic feature aggregation. Before the final output projection through a 1×1 pointwise convolution, the reordered features are restored to their original positions to maintain spatial consistency.

 \textbf{Histogram Self-
Attention}. Traditional convolutional operations utilize a fixed kernel size, which restricts the receptive field and primarily captures local information, making it less effective at modeling long-range dependencies compared to self-attention mechanisms. To overcome this limitation, a dynamic convolution technique is introduced. This approach reorganizes features along both the horizontal and vertical directions before applying the traditional convolution. After reordering, the features are recombined and processed through separable convolution in the next stage. The detailed steps of this process are outlined below:
\begin{equation}
\begin{aligned}
    &F_{1},F{2} = Split(F), F_{1}=Sort_{v}(Sort_{h}(F_{1})),
\\&F=Conv_{3\times3}^{d}(Conv_{1\times1}(Concat(F_{1},F_{2})))
\end{aligned}
\end{equation}
where $Conv_{1\times1}$ represents a $1\times1$ pointwise convolution, and  $Conv_{3\times3}$ denotes a $3\times3$ depthwise convolution. The operation $Concat$ refers to the concatenation along the channel dimension, while $Split$ refers to the operation of splitting features along the channel dimension. The notation $Sort_{i\in{h,v}}$
  indicates the sorting operation in either the horizontal or vertical direction.

By employing this approach, high-intensity and low-intensity pixels are rearranged into a structured image with aligned diagonal patterns, allowing the convolution to operate within a dynamic range. This arrangement enables the convolutional kernels to focus on preserving clean information while separately restoring degraded features.

Due to computational and memory efficiency concerns, existing Transformers typically utilize fixed window-sized attention or attention mechanisms limited to the channel dimension. However, fixed windows struggle to adaptively capture long-range dependencies, which are crucial for associating necessary features. Weather-induced degradation often results in complex patterns, where pixels containing background features or varying intensities of weather-related degradation require different levels of attention. To address this, we employ a histogram-based self-attention mechanism that categorizes spatial elements into bins, assigning different levels of attention both within and between bins. To facilitate parallel computation, we ensure that each bin contains an equal number of pixels during implementation. This approach allows the model to allocate attention more effectively based on the intensity and nature of the degradation, enhancing the restoration process.

Given the output of a dynamic convolution, we divide it into three components, denoted as Value feature $V\in R^{C\times\ H\times C}$, Query-key $F_{QK,1}$, $F_{QK,2}\in R^{2C\times\ H\times C}$, and then feed the input into two separate branches. First, we sort the sequence of $V$ and subsequently sort Query-Key according to the indices derived from the sorted $V$. The detailed procedure is as follows:
\begin{equation}
\begin{aligned}
 &V,d = Sort(R^{C\times HW}_{C\times H\times W}(V))
 \\& Q_{1},K_{1} = Split(Gather(R^{C\times HW}_{C\times H\times W}(F_{QK,1},d)))
 \\& Q_{2},K_{2} = Split(Gather(R^{C\times HW}_{C\times H\times W}(F_{QK,2},d)))
\end{aligned}
\end{equation}
Where $R^{C\times HW}_{C\times H\times W}$ refers to the process of reshaping a feature map from $R^{C\times H\times W}$ format to $R^{C\times HW}$. The term $d$ denotes the index corresponding to the sorted values, while the Gather operation is used to extract elements from a tensor according to specified indices.

We define two types of reshaping: one is based on a histogram reshaping (BHR), and the other is based on frequency histogram reshaping (FHR). the number of bins 
$B$ is fixed, with each bin containing $\frac{HW}{B}$ elements.
The second method, FHR, sets the frequency of each bin to $B$, with the number of bins being $\frac{HW}{B}$. In this manner, BHR allows us to extract large-scale information, where each bin contains a substantial number of dynamically positioned pixels, while FHR facilitates the extraction of fine-grained information, where each bin contains a smaller number of adjacent pixels with similar intensities. This process is represented by the following expression:
\begin{equation}
\begin{aligned}
    &A_{B} = softmax(\frac{R_{B}(Q_{1})R_{B}(K_{1})_{T}}{\sqrt{k}})R_{B}(V),
    \\& A_{F} = softmax(\frac{R_{F}(Q_{2})R_{F}(K_{2})_{T}}{\sqrt{k}})R_{F}(V),
    \\& A = A_{B}\odot A_{F}
\end{aligned}
\end{equation}

where $k$ is the number of heads, $R_{i\in[B,F)}$ dontes the reshaping operation of either BHR or FHR, and $A_{i\in[B,F)}$ represents the obtained attention map.

\textbf{Multi-scale Feature-fusion Feed-forward}. we established the effectiveness of a rich multi-scale representation in enhancing the performance of depth super-resolution. The MSFFN module first takes an input tensor $\mathbf{X}_{l-1} \in \mathbb{R}^{H \times W \times C}$, where $H$ is the height, $W$ is the width, and $C$ is the number of channels.
Then, the input tensor undergoes layer normalization, which helps in stabilizing the learning process by normalizing the inputs. A 1$\times$1 convolution is applied to the normalized tensor to expand the channel dimension by a factor of $r$. This expanded tensor is then fed into two parallel branches. Lastly, Combine the outputs of the two branches to create a fused tensor with a rich multi-scale representation. The entire procedure can be expressed as follows: 

\begin{align}
&\hat{\mathbf{X}}_l = f_{1 \times 1}^c(\text{LN}(\mathbf{X}_{l-1})),  \\
&\mathbf{X}_l^{p_1} = \sigma(f_{3 \times 3}^{dwc}(\hat{\mathbf{X}}_l)), \mathbf{X}_l^{s_1} = \sigma(f_{5 \times 5}^{dwc}(\hat{\mathbf{X}}_l)), \nonumber \\
&\mathbf{X}_l^{p_2} = \sigma(f_{3 \times 3}^{dwc}[\mathbf{X}_l^{p_1}, \mathbf{X}_l^{s_1}]), \mathbf{X}_l^{s_2} = \sigma(f_{5 \times 5}^{dwc}[\mathbf{X}_l^{s_1}, \mathbf{X}_l^{p_1}]), \nonumber\\
&\mathbf{X}_l = f_{1 \times 1}^c[\mathbf{X}_l^{p_2}, \mathbf{X}_l^{s_2}] + \mathbf{X}_{l-1}\nonumber 
\end{align}

where $\sigma(\cdot)$ denotes a ReLU activation function, $f_{c}^{1 \times 1}$ represents a $1 \times 1$ convolution, $f_{3 \times 3}^{dwc}$ and $f_{5 \times 5}^{dwc}$ correspond to $3 \times 3$ and $5 \times 5$ depth-wise convolutions, respectively, and $[\cdot]$ symbolizes channel-wise concatenation. 

\subsection{Adaptive Mixup For Feature Preserving}
The proposed network incorporates an encoder-decoder architecture that can effectively extract low-level features from the input image and task-specific features from the degraded image. Adaptive upsampling is utilized to enable the effective mixing of task information and image features. Addition-based skip connections, which are commonly used in encoder-decoder models, may lead to loss of shallow features or external task information. To address this, the Adaptive Mixup approach is introduced, which is able to retain more texture information of the image by adaptively mixing the features from different levels of the network. The output of Adaptive Mxiup can be expressed as:

\begin{eqnarray}
        f_{\uparrow i+1} \!\!=\!\! Mix(f_{\downarrow m-i}, f_{\uparrow i})\!\!=\!\!
        \sigma (\theta _{i})*f_{\downarrow 
i}\!+\!(1\!-\!\sigma (\theta_{i})\!*\!f_{\uparrow i})
    \end{eqnarray}
    
Where $f_{\uparrow i}$ and $f_{\downarrow m-i}$ represent the upsampling and downsampling feature maps of the i-th stage ($i\subseteq \{1,2...\\m$\}), $\sigma(\theta_{i})$ represents the learnable factor of the i-th stage, which is used to fuse the low-level features from the downsampling and the task features from the decoder, and Its value is determined by the sigmoid operator on the parameter $\theta_{i}$.

\subsection{Loss Function}
Our loss consists of $L_{smoothL_{1}}$, $L_{preceptual}$, and $L_{frequency}$. $L_{smoothL_{1}}$ is defined as follows:
\begin{eqnarray}
    L_{smoothL_{1}}=\left\{\begin{matrix}
0.5E^{2}\qquad\quad  if|E|< 1  
 \\
|E|-0.5 \qquad otherwise
\end{matrix}\right.
\end{eqnarray}
where $E=\hat{I}-G $. $\hat{I}$ is the prediction and $G$ is the ground truth.
We use the $3$rd, $8$th, and $15$th layers of VGG16 to extract features and calculate perceptual loss.
The $L_{preceptual}$ is formulated as follows:
\begin{eqnarray}
    L_{preceptual}=L_{MSE}(VGG_{3,8,15}(\hat{I} ),VGG_{3,8,15}(G))
\end{eqnarray}

Where $L_{MSE}$ denotes a mean square error loss, which can be expressed as follows:
\begin{eqnarray}
L_{MSE}(\hat{Y},Y )=\frac{1}{n} \sum_{i=1}^{n}(Y_{i}-\hat{Y_{i}})^2 
\end{eqnarray}

Where $\hat{Y}$ represents the result of network prediction, and $Y$ represents GroundTruth. $\hat{Y_i}$ and $Y_i$ represent the result of network prediction and the value of the ith pixel of GroundTruth, respectively, and $n$ represents the total number of pixels in the output image.

The $L_{frequency}$ can be summarized as:
\begin{eqnarray}
    L_{frequency}=L_{MSE}(FFT(\hat{I} ),FFT(G))
\end{eqnarray}

Where $FFT(\cdot)$ is the Fourier transform of the input.
The total loss can be summarized as:
\begin{eqnarray}
    L_{total}=L_{smoothL_{1}}+\lambda L_{perceptual}+\beta L_{frequency}
\end{eqnarray}
where $\lambda$  and $\beta$ are weights that controls the contribution of $L_{smoothL_{1}}$, $L_{frequency}$ and $L_{perceptual}$. In subsequent experiments, we have demonstrated that the network achieves the optimal fitting effect when $\lambda$ is set to 0.04 and $\beta$ is set to 0.004.

\section{Experiment}
\label{sec:experiment}

\subsection{Implementation Details}
%\textcolor{blue}{
We use the same training set as All-in-One Network\cite{RuotengLi2020AllIO} and TransWeather\cite{JeyaMariaJoseValanarasu2022TransWeatherTR} to train our network.The training set contains 9000 images of Snow100K\cite{YunFuLiu2017DesnowNetCD}, 1069 images in Raindrop\cite{RuiQian2017AttentiveGA} and 9000 Outdoor-Rain images\cite{LiRuoteng2019HeavyRI}.Sonw100K contains synthetic images with snow degradation, Raindrop contains real raindrop images, and Outdoor-Rain contains synthetic degraded images with haze stripes and rain streaks 
We test our model on Test1 dataset\cite{LiRuoteng2019HeavyRI}, RainDrop test set\cite{RuiQian2017AttentiveGA} and Snow100K-L test set\cite{YunFuLiu2017DesnowNetCD}.%}

%\textcolor{blue}{
We train our models on NVIDIA RTX3090Ti. We use the Adam optimizer and the learning rate is set to 0.0002. We use a learning rate scheduler that anneals the learning rate by 2 after 100 and 150 epochs. Our learning epoch and batch size are set to 350 and 32 respectively.%}

\section{Result}
\label{sec:result}
In this section, we conducted an extensive experimental analysis to validate the effectiveness of our proposed approach. Specifically, we provide detailed information regarding the dataset used, experimental design, and comparative analysis with state-of-the-art techniques. Furthermore, we performed various ablation experiments to verify the efficacy of our proposed module and assessed the rationality of the selected parameters.

\begin{table*}[t]
\renewcommand\arraystretch{0.7}
\caption{Quantitative Comparison on the Test1 (rain+haze) dataset based on PSNR and SSIM. DHF represents De-Hazing First and DRF represents De-Raining First. ↑ means the higher the better.
 }
\centering
\begin{tabular}{p{2cm}<{\centering}|p{5cm}<{\centering}p{3cm}<{\centering}p{2cm}<{\centering}p{2cm}<{\centering}}
\hline
\hline
Type                                                                      & Method                 & Venue    & PSNR$\uparrow$           & SSIM  $\uparrow$           \\ \hline
\multirow{8}{*}{\begin{tabular}[c]{@{}c@{}}Type  Specific\end{tabular}} & DetailsNet+Dehaze(DHF\cite{Fu_Huang_Zeng_Huang_Ding_Paisley_2017}) & CVPR2017 & 13.36          & 0.5830          \\
                                                                          & DetailsNet+Dehaze(DRF\cite{Fu_Huang_Zeng_Huang_Ding_Paisley_2017}) & CVPR2017 & 15.68          & 0.6400          \\
                                                                          & RESCAN+Dehaze(DHF\cite{XiaLi2018RecurrentSC})     & CVPR2018 & 14.72          & 0.5870          \\
                                                                          & RESCAN+Dehaze(DHF\cite{XiaLi2018RecurrentSC})     & CVPR2018 & 15.91          & 0.6150          \\
                                                                          & pix2pix\cite{PhillipIsola2016ImagetoImageTW}                & CVPR2017 & 19.09          & 0.7100          \\
                                                                          & HRGAN\cite{LiRuoteng2019HeavyRI}                  & CVPR2019 & 21.56          & 0.8550          \\
                                                                          & Swin-IR\cite{JingyunLiang2021SwinIRIR}                & CVPR2021 & 23.23          & 0.8685          \\
                                                                          & MPRNet\cite{SyedWaqasZamir2021MultiStagePI}                 & CVPR2021 & 21.29          & 0.8456          \\ \hline
\multirow{6}{*}{\begin{tabular}[c]{@{}c@{}}Multi-Task\end{tabular}}                                               & All-in-One\cite{RuotengLi2020AllIO}             & CVPR2020 & 24.71          & 0.8980          \\
                                                                          & TransWeather\cite{JeyaMariaJoseValanarasu2022TransWeatherTR}           & CVPR2022 & 26.69          & 0.9501          \\
                                                                          & Zhen's\cite{Tan_Wu_Liu_Chu_Lu_Ye_Yu_2023} &TIP2023& \textbf{30.02}&0.9078\\
                                                                          & AIRFormer\cite{Gao_Wen_Zhang_Zhang_Chen_Liu_Luo_2024} &TCSVT2023& \textbf{29.23} &0.9350\\ & TTAMS\cite{wen2024multipleweatherimagesrestoration}                   & CGI2024      & 27.61 & 0.9595
                                                                     \\ & Ours                   & --       & 27.64 & \textbf{0.9598}     \\ \hline
                                                                          \hline
\end{tabular}
\label{table1}
\end{table*}

\begin{table*}[t]
\renewcommand\arraystretch{0.7}
\caption{Quantitative Comparison on the SnowTest100k-L dataset based on PSNR and SSIM.  ↑ means the higher the better.
 }
 \centering
\begin{tabular}{p{2cm}<{\centering}|p{5cm}<{\centering}p{3cm}<{\centering}p{2cm}<{\centering}p{2cm}<{\centering}}
\hline
\hline
Type                                                                      & Method                 & Venue    & PSNR$\uparrow$            & SSIM$\uparrow$             \\ \hline
\multirow{5}{*}{\begin{tabular}[c]{@{}c@{}}Type  Specific\end{tabular}} &  pix2pix\cite{PhillipIsola2016ImagetoImageTW}                          & CVPR2017 & 28.02                                 & 0.8547                                 \\
                                                                           & Attn.GAN\cite{RuiQian2017AttentiveGA} & CVPR2018 & 30.55                                 & 0.9023                                 \\
                                                                           & Quan et al\cite{YuhuiQuan2019DeepLF}                       & ICCV2019 & 31.44                                 & 0.9263                                 \\
                                                                           & Swin-IR\cite{JingyunLiang2021SwinIRIR}                          & CVPR2021 & 30.82                                 & 0.9035                                 \\ \hline
\multirow{6}{*}{\begin{tabular}[c]{@{}c@{}}Multi-Task\end{tabular}}                                               & All-in-One\cite{RuotengLi2020AllIO}             & CVPR2020 & 28.33          & 0.8820          \\
 & TransWeather\cite{JeyaMariaJoseValanarasu2022TransWeatherTR}           & CVPR2022 & 28.02          & 0.9305          \\& Zhen's\cite{Tan_Wu_Liu_Chu_Lu_Ye_Yu_2023} &TIP2023& \textbf{29.78}&0.8907\\
                                                                          & AIRFormer\cite{Gao_Wen_Zhang_Zhang_Chen_Liu_Luo_2024} &TCSVT2023& 29.15 &0.9270
                                                                          \\ 
  & TTAMS\cite{wen2024multipleweatherimagesrestoration}                   & CGI2024       & 28.48 & 0.9367
  \\
  & Ours                   & -       & 28.70 & \textbf{0.9382}
  \\ \hline
\hline
\end{tabular}
\label{table2}
\end{table*}

\begin{table*}[t]
\renewcommand\arraystretch{0.7}
\caption{Quantitative comparison on the RainDrop test dataset based on PSNR and SSIM. ↑ means the higher the better.
 }
 \centering
\begin{tabular}{p{2cm}<{\centering}|p{5cm}<{\centering}p{3cm}<{\centering}p{2cm}<{\centering}p{2cm}<{\centering}}
\hline
\hline
Type                                                                      & Method                 & Venue    & PSNR$\uparrow$            & SSIM$\uparrow$             \\ \hline
\multirow{5}{*}{\begin{tabular}[c]{@{}c@{}}Type  Specific\end{tabular}} &  pix2pix\cite{PhillipIsola2016ImagetoImageTW}                          & CVPR2017 & 28.02                                 & 0.8547                                 \\
                                                                           & Attn.GAN\cite{RuiQian2017AttentiveGA} & CVPR2018 & 30.55                                 & 0.9023                                 \\
                                                                           & Quan et al\cite{YuhuiQuan2019DeepLF}                       & ICCV2019 & 31.44                                 & 0.9263                                 \\
                                                                           & Swin-IR\cite{JingyunLiang2021SwinIRIR}                          & CVPR2021 & 30.82                                 & 0.9035                                \\ \hline
\multirow{6}{*}{\begin{tabular}[c]{@{}c@{}}Multi-Task\end{tabular}}                                               & All-in-One\cite{RuotengLi2020AllIO}             & CVPR2020 & 31.12          & 0.9268          \\
 & TransWeather\cite{JeyaMariaJoseValanarasu2022TransWeatherTR}           & CVPR2022 & 28.84          & 0.9527          \\& Zhen's\cite{Tan_Wu_Liu_Chu_Lu_Ye_Yu_2023} &TIP2023& 31.03&0.9228\\
                                                                          & AIRFormer\cite{Gao_Wen_Zhang_Zhang_Chen_Liu_Luo_2024} &TCSVT2023& \textbf{32.09} &0.9450
                                                                          \\ 
  & TTAMS\cite{wen2024multipleweatherimagesrestoration}                   & CGI2024       & 29.35 & 0.9574
  \\ 
  & Ours                   & --       & 29.40 & \textbf{0.9587}
  \\ \hline
\hline
\end{tabular}
\label{table3}
\end{table*}

 \begin{figure*}
\begin{center}
    \includegraphics [width=1\textwidth]{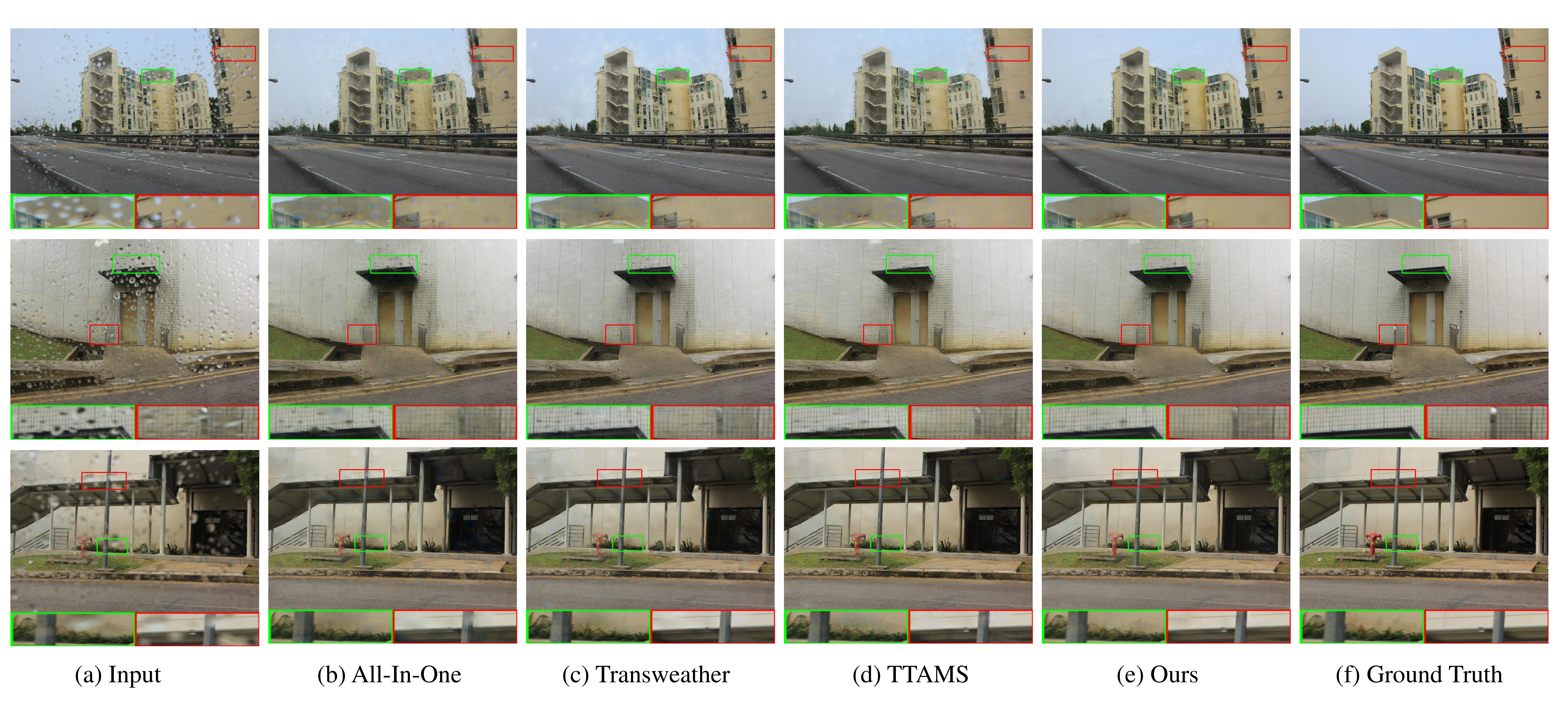}
    \end{center}
\caption {Qualitative results comparison of the proposed method with existing state-of-the-art methods All-in-One\cite{RuiQian2017AttentiveGA} , TransWeather\cite{JeyaMariaJoseValanarasu2022TransWeatherTR}, and TTAMS\cite{wen2024multipleweatherimagesrestoration} on Raindrop dataset\cite{RuiQian2017AttentiveGA} for raindrop removal.}
\label{figure6}
\end{figure*}

 \begin{figure*}
\begin{center}
    \includegraphics [width=1\textwidth]{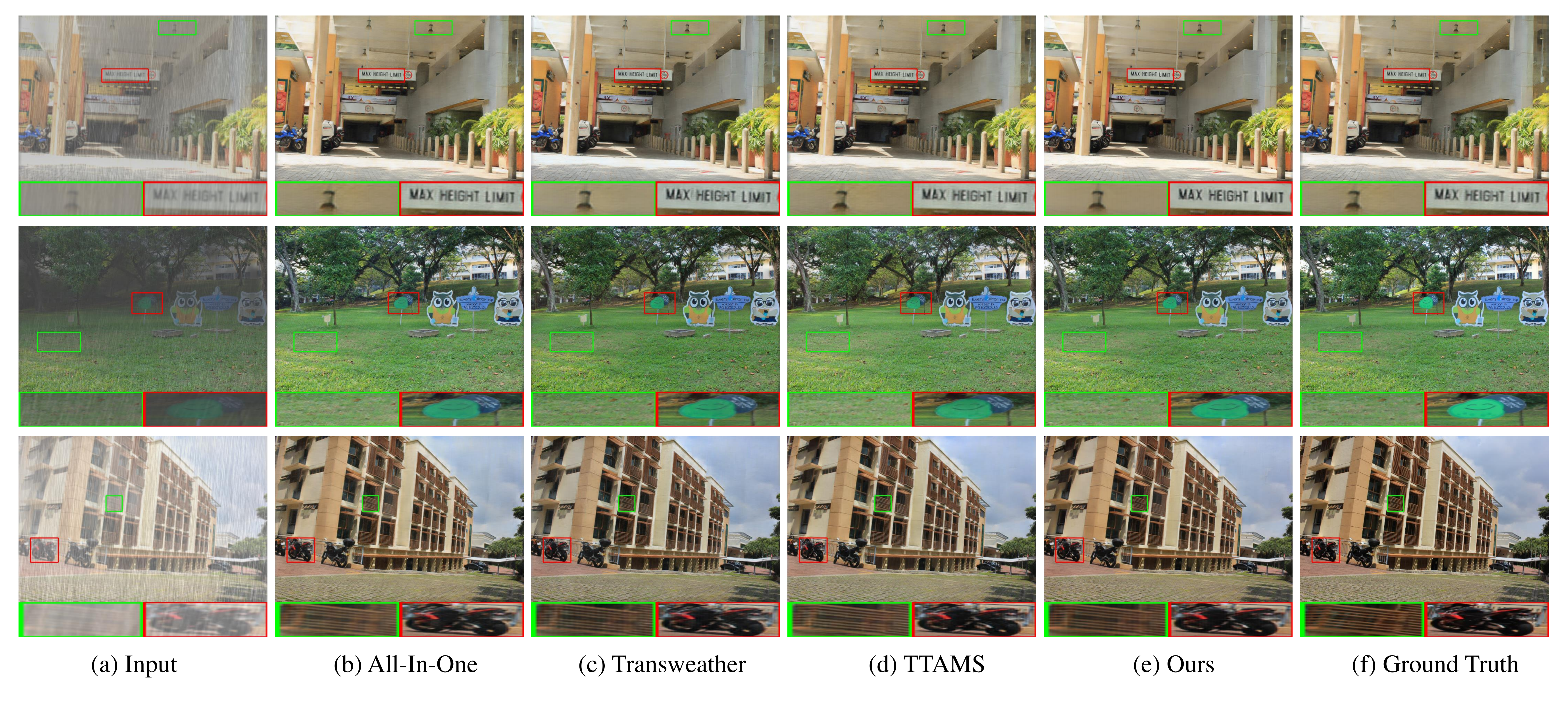}
    \end{center}
\caption {Qualitative results comparison of the proposed method with existing state-of-the-art methods All-in-One\cite{RuiQian2017AttentiveGA} , TransWeather\cite{JeyaMariaJoseValanarasu2022TransWeatherTR} and TTAMS\cite{wen2024multipleweatherimagesrestoration}, on Test1 dataset\cite{LiRuoteng2019HeavyRI} for rain removal and haze removal.}
\label{figure7}
\end{figure*}

 \begin{figure*}
\begin{center}
    \includegraphics [width=1\textwidth]{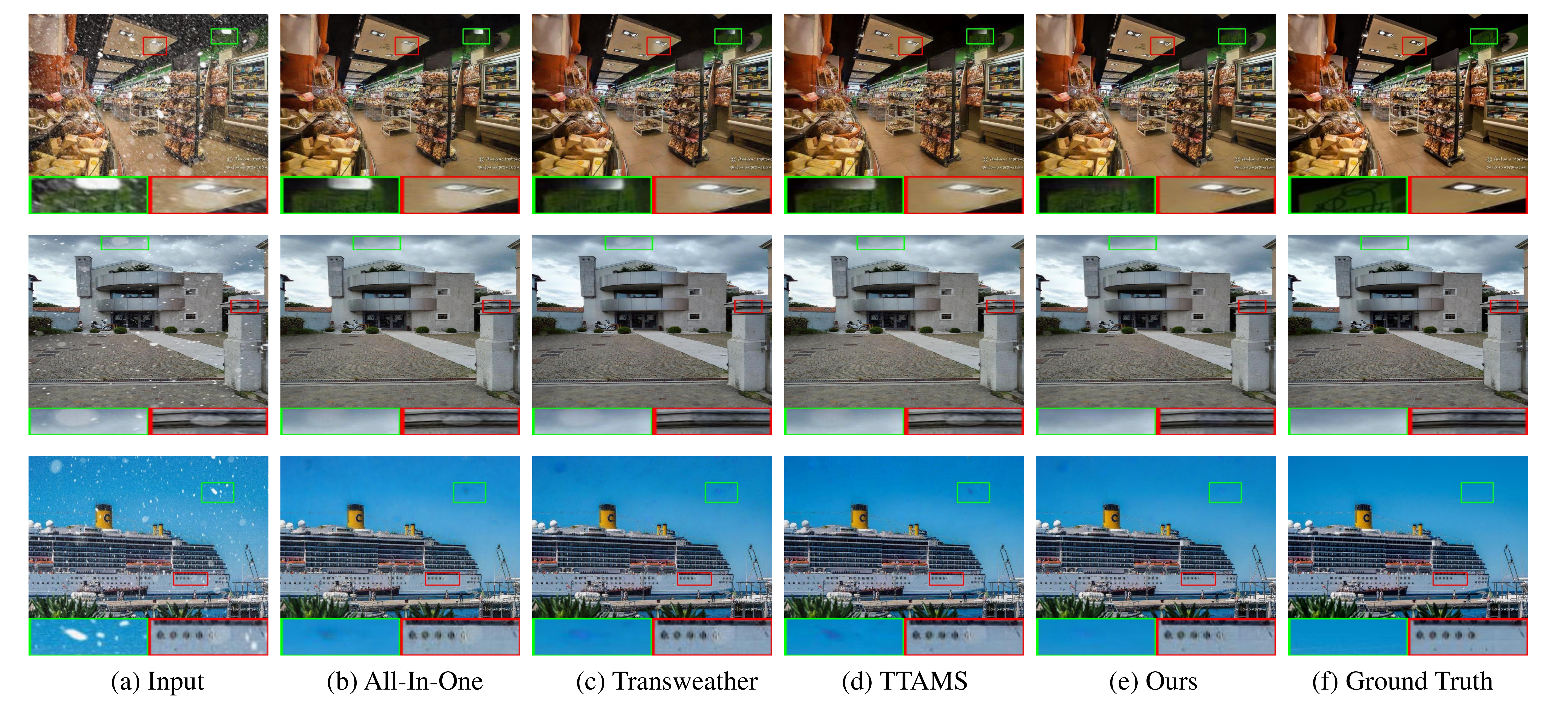}
    \end{center}
\caption {Qualitative results comparison of the proposed method with existing state-of-the-art methods All-in-One\cite{RuotengLi2020AllIO} , TransWeather\cite{JeyaMariaJoseValanarasu2022TransWeatherTR}, and TTAMS\cite{wen2024multipleweatherimagesrestoration} on Snow100K dataset\cite{YunFuLiu2017DesnowNetCD} for snow removal.}
\label{figure8}
\end{figure*}

\subsection{Comparison With State-of-The-Art}
We compare our method against state-of-the-art modalities specifically designed for each task. For the raindrop task, we compare with state-of-the-art methods such as Attention Gan\cite{RuiQian2017AttentiveGA} and complementary cascaded network(CCN)\cite{RuijieQuan2021RemovingRA}. For the snow task, we compare with state-of-the-art methods such as JSTASR\cite{ChenWeiTing2020JSTASRJS}, Desnow-Net\cite{YunFuLiu2017DesnowNetCD}, etc. For the rain+haze task, we compare with state-of-the-art methods such as HRGAN\cite{LiRuoteng2019HeavyRI}, Swin-IR\cite{JingyunLiang2021SwinIRIR}, etc.
At the same time, we also compare with the state-of-the-art methods of multi-tasking. The detailed experimental results will be shown below.

\subsubsection{Referenced Quality Metrics}
We use PSNR and SSIM to quantitatively evaluate the performance of different models in Test1, Snow100K-L, and RainDrop test sets. All the experimental results are shown in Tables \ref{table1}, \ref{table2} and \ref{table3}, and the test sets they respectively represent are Test1, Snow100K-L, and RainDop test sets. Our results demonstrate that our proposed method outperforms the multi-task single-job approach in the combination of three distinct weather types. Moreover, when compared to other multi-task models, our approach exhibits superior performance. 

\subsubsection{Visual Quality Comparison}
\begin{figure*}
\begin{center}
    \includegraphics [width=1\textwidth]{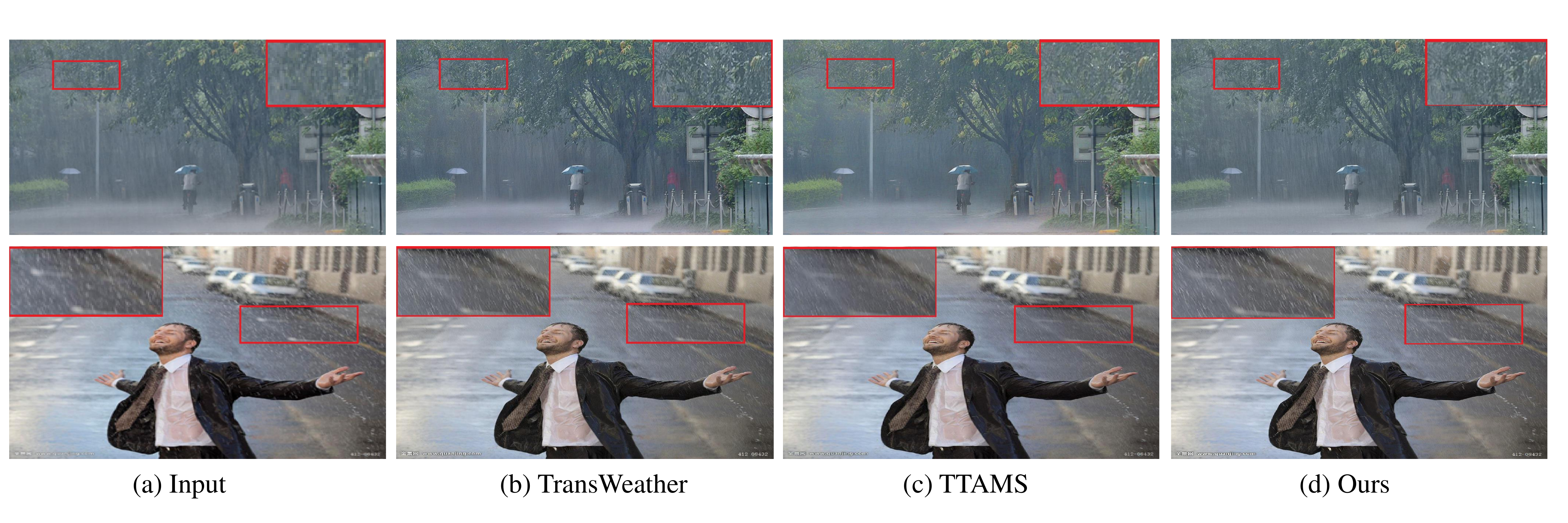}
    \end{center}
\caption {Comparing real-world severe weather image restoration examples using TransWeather\cite{JeyaMariaJoseValanarasu2022TransWeatherTR} and our method. In the above example, in the repair of leaves, our method can repair more details than TransWeather, and the interlacing of leaves is clearer. In the example below, we have better results for dense rain marks repair.}
\label{figure11}
\end{figure*}

We performed a qualitative comparison with All-in-One Network and TransWeather. The results are shown in Figures 
 \ref{figure6}, Figures \ref{figure7}, and Figures \ref{figure8}. Our proposed method exhibits superior performance in removing degradations, particularly in cases where there are large areas of degenerate features. Furthermore, our approach outperforms other models in terms of restoring texture and color information in the image.  We also conducted evaluations using real-world image restoration test cases to assess the performance of models trained on synthetic data. In order to provide visual comparisons, we compared our model with the recent TransWeather network, both of which are specialized in multi-weather recovery. Figure\ref{figure11} showcases a qualitative rain removal comparison between selected images from various real rain image sets, namely Rain Mist\cite{Li_Araujo_Ren_Wang_Tokuda_Junior_Cesar-Junior_Zhang_Guo_Cao_2019}. In the first example shown in Figure\ref{figure11}, our model demonstrates superior restoration capabilities in terms of preserving complex textures and details, particularly evident in the restoration of leaves. In the second example, our model exhibits a cleaner overall effect in repairing rain marks compared to TransWeather and TTAMS.

% \begin{table}[H]
% \renewcommand\arraystretch{0.7}
% \caption{Ablation Study on the Test1 test dataset.
% }
% \scalebox{0.59}{
% \begin{tabular}{c|c|c}
% \hline
% \hline
% Method                           & PSNR(↑)                       & SSIM(↑)                        \\ \hline
% Base                             & 26.69                         & 0.9501 \\
% transformerIPB                       & 26.92 & 0.9518 \\
% transformerIPB+Task sequence generator                & 27.16                        & 0.9539                         \\
% TIPB+Task sequence generator + Histogram Transformer  & 27.50        & 0.9568   \\ 
% TIPB+Task sequence generator +Histogram Transformer  +  Adaptive Mixup  & \textbf{27.64}                         & \textbf{0.9598}\\

% \hline
% \hline
% \end{tabular}}
% \label{table4}
% \end{table}

% \usepackage{booktabs}

\begin{table}[]
 \caption{Ablation Study on the Test1 test dataset.
 }
 \resizebox{\linewidth}{!}{
\begin{tabular}{@{}ccccccc@{}}
\toprule
Methods & Base & TIPB & TSG & Histogram Transformer & Adaptive Mixup & PSNR/SSIM             \\ \midrule
(a)     & \checkmark    &      &     &                       &                & 26.69/0.9501          \\
(b)     & \checkmark    & \checkmark    &     &                       &                & 26.92/0.9518          \\
(c)     & \checkmark    & \checkmark    & \checkmark   &                       &                & 27.16/0.9539          \\
(d)     & \checkmark    & \checkmark    & \checkmark   & \checkmark                     &                & 27.50/0.9568          \\
(e)     & \checkmark    & \checkmark    & \checkmark   & \checkmark                     & \checkmark              & \textbf{27.64/0.9598} \\ \bottomrule
\end{tabular}
}
\label{table4}
\end{table}

\subsubsection{Object Detection Comparison}
Severe weather conditions significantly impact the field of autonomous driving, particularly in the context of object detection from acquired images. The ability to accurately detect objects in these images is crucial for analyzing the current driving situation and making appropriate judgments. In this chapter, we utilize the YOLOV5 algorithm to perform object detection on images repaired by each of the considered models.

\begin{figure*}
\begin{center}
    \includegraphics [width=1\textwidth]{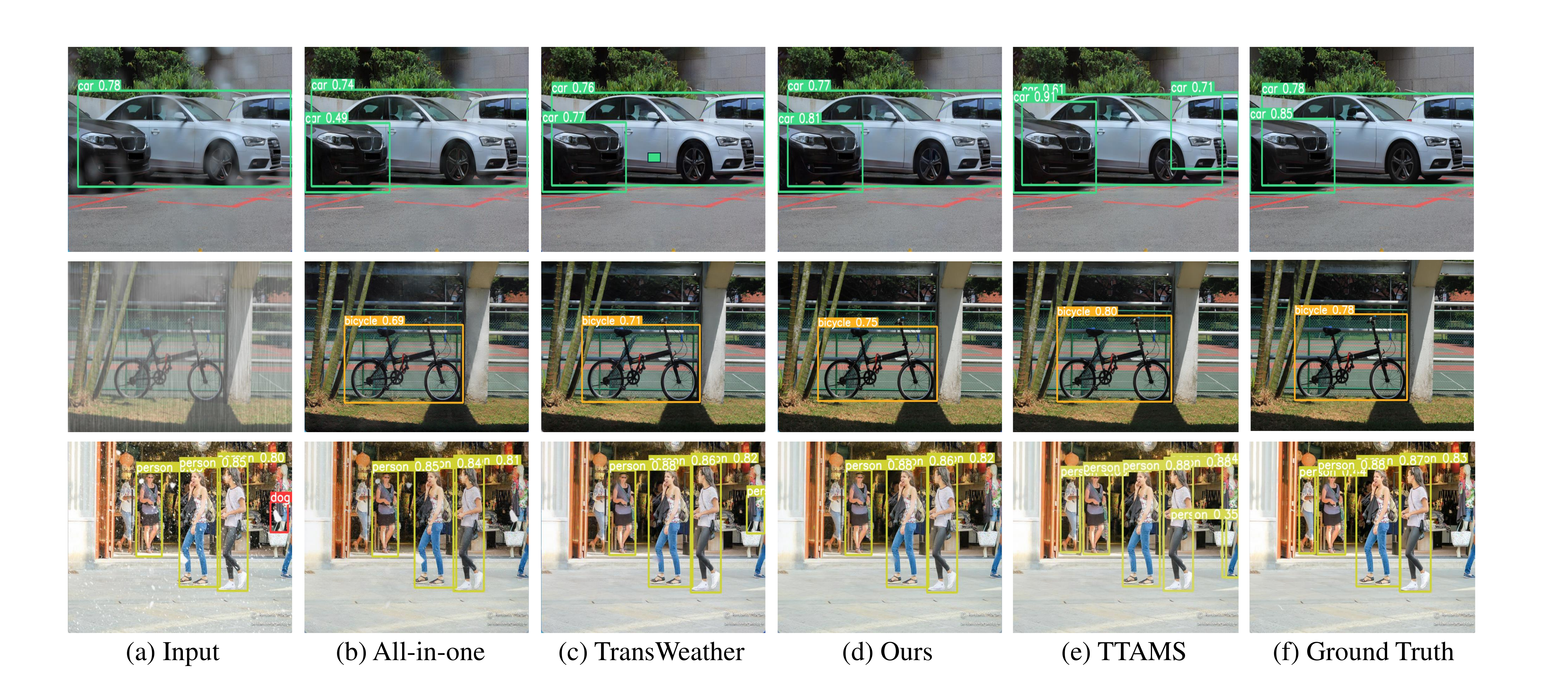}
    \end{center}
\caption {Object Detection Comparison of the proposed method with existing state-of-the-art methods All-in-One\cite{RuiQian2017AttentiveGA} and TransWeather\cite{JeyaMariaJoseValanarasu2022TransWeatherTR}, Using YOLOv5 for object detection. What is framed is the recognized object, and above the frame is the type and confidence of the recognized object. Compared with TransWeather, our model has higher accuracy and confidence.}
\label{figure10}
\end{figure*}

Figure \ref{figure10} illustrates that our model exhibits superior confidence in object detection compared to the All-in-One\cite{RuotengLi2020AllIO} , TransWeather\cite{JeyaMariaJoseValanarasu2022TransWeatherTR} and TTAMS\cite{wen2024multipleweatherimagesrestoration} models. Moreover, TransWeather even demonstrates an erroneous judgment in the final scene, posing a significant danger to autonomous driving. Consequently, our model proves more suitable for deployment and implementation in this domain. Table\ref{table5} presents the quantitative analysis outcomes of our object detection methodology on a dataset comprising 200 objects. In terms of detection accuracy, our approach surpasses the performance of the All-in-One\cite{RuotengLi2020AllIO} method. Furthermore, when compared to the TransWeather\cite{JeyaMariaJoseValanarasu2022TransWeatherTR} technique, our detected objects exhibit a higher level of confidence.

\begin{table}[H]
\caption{Quantitative comparison of object detection with TransWeather , All-in-one and TTAMS. Errors and Omissions are the sum of the number of false detections and missed detections. Average Confidence is the average confidence of the recognized objects.}
\scalebox{0.9}{
\begin{tabular}{c|c|c}
\hline
\hline
             & Errors and Omissions & Average Confidence \\ \hline
Input        & 144                  & 0.184              \\
All-In-One\cite{RuotengLi2020AllIO}   & 18                   & 0.624              \\
TransWeather\cite{JeyaMariaJoseValanarasu2022TransWeatherTR} & 3                    & 0.648              \\
TTAMS\cite{wen2024multipleweatherimagesrestoration}         & 3                    & 0.659              \\
Ours         & 0                    & 0.668              \\
GroundTruth  & 0                    & 0.681\\
\hline
\hline
\end{tabular}}
\label{table5}
\end{table}

\subsection{Ablation Study}
In this section, we conduct an evaluation of the contribution of each component in our methods. To do so, we progressively replace the Base model with our components and train different networks in the same environment. We then calculate the average PSNR and SSIM in the Test1 dataset and present the results in Table \ref{table4}. Through this evaluation, we demonstrate the effectiveness of each component of our approach.

\section{Conclusion}
\label{sec:conclusion}
In this paper, we propose a novel model to address the challenges posed by multi-weather degraded images. Our approach involves leveraging a trainable sequence to extract multi-scale features, which are subsequently used to generate task sequences specific to degradation-related tasks. These task sequences guide the network, enabling it to selectively focus on degradation information from different tasks. To more effectively capture global and local dynamic range features, we introduce the Histogram Transformer Block. This module effectively captures long-range dependencies, aiding the restoration process. Additionally, we employ adaptive fusion to merge features obtained from different modules, thereby enhancing the overall performance of our method. Our approach overcomes the limitations of single-task-specific networks, which are often difficult to deploy in practice. Compared to other multi-task processing networks, our method demonstrates exceptional capability in extracting various weather degradation information while effectively restoring heavily degraded content. To validate the effectiveness of our proposed method, we conducted extensive evaluations on different datasets. The experimental results demonstrate the superior performance of our method, surpassing many state-of-the-art techniques in the field. 

% Note: If you are using BibTeX, please use the following code:
\bibliographystyle{unsrt}
\bibliography{bib_CGIconf}

\end{sloppypar}
\end{document}